\documentclass{article}
\usepackage[a4paper, total={6in, 8in}]{geometry}
\usepackage[utf8]{inputenc}
\usepackage[english]{babel}
\usepackage{csquotes}
\usepackage{microtype}
\usepackage{amsmath,amsthm,amssymb}
\usepackage[bookmarksdepth=3,linktoc=all,colorlinks=true,urlcolor=blue,linkcolor=blue,citecolor=blue]{hyperref}
\usepackage[capitalize]{cleveref}
\usepackage[style=ieee, sorting=none]{biblatex}

\usepackage[inkscapelatex=false]{svg}
\usepackage{siunitx}
\usepackage{algorithm}
\usepackage{algpseudocode}
\algnewcommand{\LineComment}[1]{\State \(\triangleright\) #1}
\def\b#1{\boldsymbol{#1}}
\usepackage{subcaption}

\usepackage{booktabs}

\usepackage{listings}
\usepackage{xcolor}
\usepackage{soul}

\definecolor{codegreen}{rgb}{0,0.6,0}
\definecolor{codegray}{rgb}{0.5,0.5,0.5}
\definecolor{codepurple}{rgb}{0.58,0,0.82}
\definecolor{backcolour}{rgb}{0.95,0.95,0.92}

\lstdefinestyle{mystyle}{
    backgroundcolor=\color{backcolour},   
    commentstyle=\color{codegreen},
    keywordstyle=\color{magenta},
    numberstyle=\tiny\color{codegray},
    stringstyle=\color{codepurple},
    basicstyle=\ttfamily\footnotesize,
    breakatwhitespace=false,         
    breaklines=true,                 
    captionpos=b,                    
    keepspaces=true,                 
    numbers=left,                    
    numbersep=5pt,                  
    showspaces=false,                
    showstringspaces=false,
    showtabs=false,                  
    tabsize=2
}
\usepackage{array}
\newcolumntype{H}{>{\setbox0=\hbox\bgroup}c<{\egroup}@{}}

\usepackage[most]{tcolorbox}

\begin{document}

\title{In-Situ Soil-Property Estimation and Bayesian Mapping with a Simulated Compact Track Loader}
\author{W. Jacob Wagner, Ahmet Soylemezoglu, Katherine Driggs-Campbell}
\maketitle

\begin{abstract}
Existing earthmoving autonomy is largely confined to highly controlled and well-characterized environments due to the complexity of vehicle-terrain interaction dynamics and the partial observability of the terrain resulting from unknown and spatially varying soil conditions. In this chapter, a a soil-property mapping system is proposed to extend the environmental state, in order to overcome these restrictions and facilitate development of more robust autonomous earthmoving.

A GPU accelerated elevation mapping system is extended to incorporate a blind mapping component which traces the movement of the blade through the terrain to displace and erode intersected soil, enabling separately tracking undisturbed and disturbed soil. Each interaction is approximated as a flat blade moving through a locally homogeneous soil, enabling modeling of cutting forces using the fundamental equation of earthmoving (FEE). Building upon our prior work on in situ soil-property estimation, a method is devised to extract approximate geometric parameters of the model given the uneven terrain, and an improved physics infused neural network (PINN) model is developed to predict soil properties and uncertainties of these estimates. 

A simulation of a compact track loader (CTL) with a blade attachment is used to collect data to train the PINN model. Post-training, the model is leveraged online by the mapping system to track soil property estimates spatially as separate layers in the map, with updates being performed in a Bayesian manner. Initial experiments show that the system accurately highlights regions requiring higher relative interaction forces, indicating the promise of this approach in enabling soil-aware planning for autonomous terrain shaping. 
\end{abstract}

\label{ch:istvs25}
\section{Introduction} 
The construction equipment manufacturing industry has been advancing equipment automation over the last few decades \cite{Saidi2016}.
Most earthmoving equipment can now either be retrofitted with or come directly from the factory with advanced operator assist and machine control capabilities \cite{Jackson2017,Ries2019}.
The most advanced versions of this technology not only provide feedback to the operator about the state of the current terrain compared to the design surface, but also automate portions of the control to increase inexperienced operator efficiency and to ensure that soil is not unnecessarily disturbed \cite{Azar2017, Hayashi2013}.

Given a desire to increase machine productivity, the growing labor shortage for skilled equipment operators, and advances in self-driving, there is recent increased interest in the development of autonomous construction equipment \cite{Nguyen2023,Mathur2024}.
In the domain of autonomous earthmoving, heavy equipment original equipment manufacturers (OEMs) have focused on developing of autonomy for large scale mining operations, including autonomous dump truck navigation \cite{Bellamy2011} and execution of slot-dozing operations autonomously \cite{DAdamo2023}.
Several startups have also entered the market, and are working to bring autonomy to other heavy equipment, including excavators, and to other applications such as solar-farm construction and earthmoving for construction.
For a more comprehensive treatment of autonomous construction see these reviews \cite{Nguyen2023,Dadhich2016,Ha2019}.

While in some applications, such as surface mining, it is possible to have a significant degree of control over the environmental conditions \cite{DAdamo2023}, in many applications of interest, the environment is much more unstructured \cite{Soylemezoglu2021,Zhu2023}.
The environment is partially observable because existing sensors are only capable of producing information regarding the surface characteristics of the terrain.
The shape of the terrain can be provided \textit{a priori} or mapped by inferring ground height from machine position and/or using onboard depth sensing \cite{Jud2019, DAdamo2023}.
Semantic segmentation methods can provide some information regarding the broad class of material on the surface, but provide limited information regarding the physical properties of the soil relevant to earthmoving.
Given that soil conditions can substantially vary with both lateral position and depth, surface measurements are not adequate to inform soil-aware planning and tool control.

It is generally recognized that some level of adaptation is necessary at the control level for autonomous earthmoving to enable robust and efficient execution of earthmoving-cycles in environments with unknown and spatially varying soil conditions \cite{Dadhich2016}.
However, adaptation of earthmoving plans to soil-conditions has not been thoroughly investigated.

\begin{figure}[h!]
    \centering
    \includegraphics[width=0.75\linewidth]{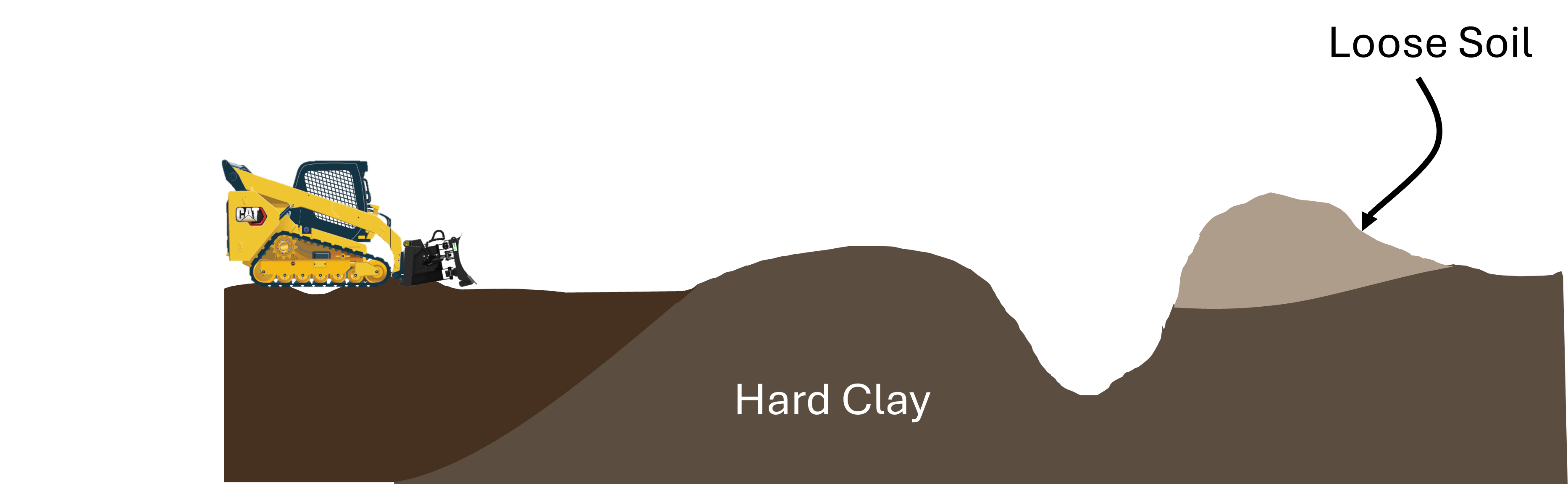}
    \caption{Example earthmoving scenario used to illustrate the need for soil-aware planning}
    \label{fig:ditch}
\end{figure}

To illustrate the need for adaptation to soil conditions at the planning level, consider the following scenario, depicted in Figure \ref{fig:ditch}.
The task is for a bulldozer to fill in a ditch and ensure the resulting terrain surface smooth enough for a wheeled vehicle to traverse the region.
The planner's job is to select a cut start location and a desired relative cut-depth for the blade controller.
It is important to note that the soil strength, arising from the soil properties, determines the maximum allowable cut depth to avoid machine stalling given a finite bulldozer tractive force.
If the soil properties are unknown it is impossible to select the dig parameters optimally.
A soil-ignorant planner can act conservatively by selecting a cut start location far from the deposit point and a small depth of cut, but this is only optimal for very high-strength soils.
Alternatively, the planner can act more aggressively by selecting a dig point near the deposit location and a large depth of cut.
This approach is optimal for uniform low-strength soils, but leads to insufficient accumulated material volume when an adaptive low-level blade controller is utilized to avoid stalling.
In fact, this scenario is encountered by Egli et. al. in the development of a learned adaptive scooping policy, where it is remarked that for hard soils the controller is only able to partially fill the bucket, even when the user defined dig-point is selected to be at the maximum extent of the excavator arm \cite{Egli2022}.

Soil conditions are often not spatially homogeneous, which can further complicate planning.
Consider a slightly modified scenario where far from the ditch, a low-strength loam is present, but near the ditch the soil is made of high-strength clay.
If a soil-ignorant planner is tuned aggressively to make deep cuts near the ditch, this will result in a much higher number of cuts to fill the ditch than a soil-aware planner which can more quickly accumulate material by choosing to cut the lower-strength material further away from the ditch and then carry it over the higher-strength material to the deposit location.
The possibly obvious but important lesson here is that spatially varying soil conditions result in varying forces required to shear soil, which subsequently affect earthmoving plan feasibility and the resulting cut trajectory, further affecting accumulation and deposition of material.
The implication is that in order to support development of soil-aware earthmoving planners, it is first necessary to develop an approach to obtaining knowledge of soil conditions in situ and to devise a method for keeping track of how these conditions vary spatially, which is the focus of this work.




\subsection{Soil Aware Autonomous Earthmoving}
Early approaches to autonomous earthmoving focused on the development of low-level control systems to coordinate machine movement throughout the phases of a dig cycle.
One approach is to formulate the control problem in terms of kinematic trajectory tracking.
This is intuitive, especially for excavation applications, given that an excavator arm is a serial manipulator and can be treated similarly to robotic manipulators, enabling re-use of motion planning tools for trajectory generation and task-space control.
However, these kinematic trajectory control approaches fail if the provided trajectory is unable to be executed accurately when machine force limits are reached.
The force required to move the tool through the ground depends not only on the dynamics of the machine, but on the resistance of the soil to shearing, which is dependent on the soil properties.

Rule-based methods and fuzzy logic controllers were initially explored to enable adaptation to soil conditions, but require elicitation of rules from expert operators which can be challenging and have fallen out of favor recently \cite{Bradley1998, Lever2001}. 
Another approach is to use compliant control to balance the trajectory tracking objective with machine force constraints \cite{Dobson2016, Ha2000a} 
A challenge with using these compliant methods, impedance control for example, is that the impedance of the environment is unknown and variable.
If a fixed environmental impedance is assumed, then tracking performance will vary significantly across soil types.
The open loop bandwidth of hydraulic systems on heavy equipment is limited and high gains can excite low resonant modes of the tool leading to dangerous oscillations \cite{Maeda2015}.
One way of addressing this tuning problem is to adapt controller parameters, e.g. stiffness and damping, online \cite{Feng2022,Ha2000a}.
Sotiropoulos and Asada propose an alternative compliant control formulation based on maximization of power transmitted to the soil, that aims to match the impedance of the excavator with the impedance of the load, but it is not clear how this approach can be extend for tasks that require achieving a desired terrain shape \cite{Sotiropoulos2019}.
Maeda et. al. suggest taking advantage of the cyclical nature of the excavation dig-cycles by using iterative learning control to compensate for the interaction forces encountered in the subsequent digging pass \cite{Maeda2015}.
Jud et. al. discards the idea of tracking a desired kinematic trajectory and instead details a force trajectory control approach to digging that supports constraining positions, velocities, and forces while avoiding penetration of a design surface \cite{Jud2017}.
This scheme results in the production of various kinematic trajectories as the system reacts to differing soil conditions, but significant retrofitting of the machine hydraulic system is necessary to attain sufficient force tracking performance \cite{Jud2019}.

More conventional compliant manipulation approaches can also be used successfully, and can be combined with methods that predict interaction forces and/or disturbances to improve tracking and ensure selection of feasible trajectories \cite{Maeda2015,Singh1995a,Mononen2022a}.
The main idea is that during earthmoving, when a system deviates from the model-derived expected behavior, this is caused by unmodeled soil-tool interaction forces and therefore, tracking errors actually contain information regarding the soil conditions.
These errors are evident at the force level, but are also reflected at the trajectory and velocity level \cite{Bradley1995,Maeda2013}, which helps to explain how human operators can infer information about soil conditions without knowledge of interaction forces.

Rather than treating interaction forces as disturbances, they can be predicted and incorporated in a feed-forward manner to enhance control \cite{Cannon2000}.
Singh’s pioneering work introduced the concept of soil property identification to enable adaptive tactical planning, allowing for efficient excavation while staying within machine force limits \cite{Singh1995a}. By estimating soil properties iteratively, it becomes possible to optimize digging trajectories under force constraints, thereby preventing actuator saturation. These properties are inferred from the soil-tool interaction using a combination of terrain mapping, known tool trajectories, and force measurements collected during execution \cite{Luengo1998,Singh1995,Cannon1999}. For a comprehensive review of inverse terramechanics methods—encompassing both traditional techniques and machine learning approaches—see Arreguin et al. \cite{Arreguin2021}.

\begin{figure}[h!]
    \centering
    \includegraphics[width=0.5\linewidth]{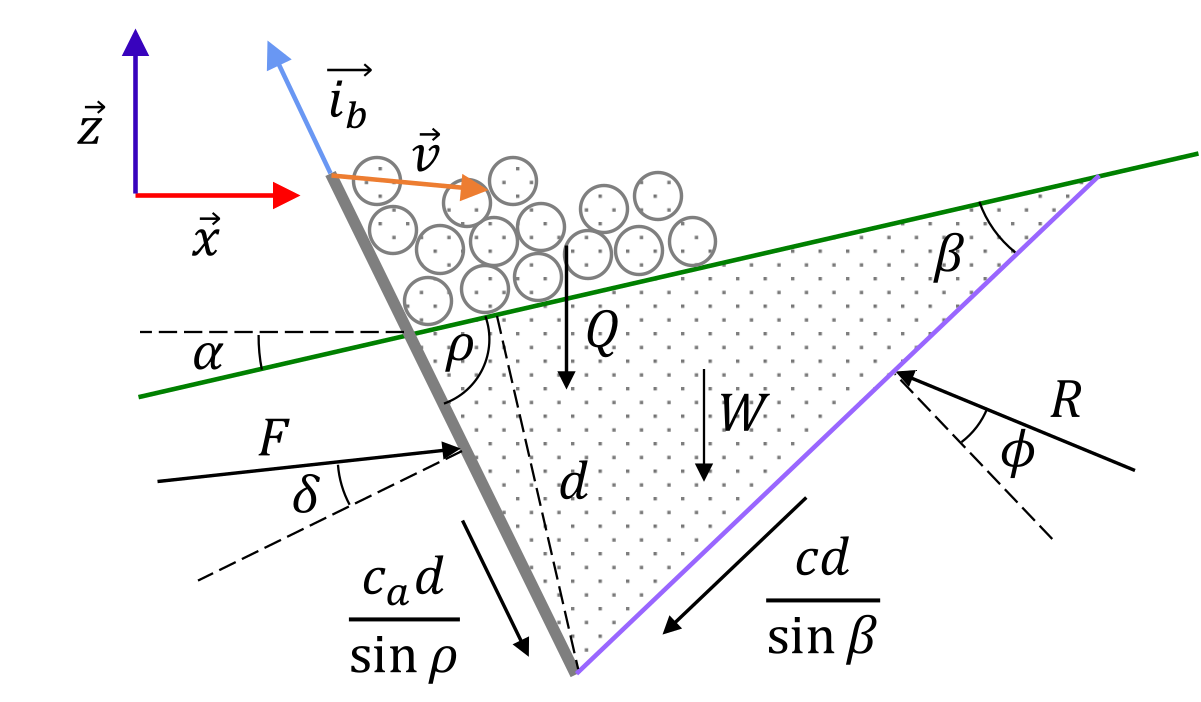}
    \caption{Force diagram for the fundamental equation of earthmoving (FEE) with a surface inclined at angle $\alpha$}
    \label{fig:FEE}
\end{figure}

Many semi-empirical models of soil-tool interaction exist with varying degrees of complexity and different parametrizations, but the fundamental equation of earthmoving (FEE) is a common choice \cite{Blouin2001}.
This model is derived by assuming a Mohr-Coulomb model of soil shear strength and performing a force balance on a system with a flat angled blade failing a wedge of soil \cite{mckyes2012agricultural,Reece1964,Holz2013, Wagner2025FEE}, as depicted in Figure \ref{fig:FEE}.
The resulting basic equation is given by
\begin{align} \label{eq:FEE_v2}
    f = f_{\text{FEE}}(\Theta) &= \gamma d^2 w N_\gamma + c d w N_c + Q N_Q + c_a d w N_a \\
    F(\Theta) &= f \cdot \left[ \sin(\rho+\delta-\alpha),\ \cos(\rho+\delta-\alpha)\right] \\
    &\hspace*{-4em}\begin{array}{cccc}  \tag{2.3--2.6} 
        N_\gamma = \dfrac{[\cot(\rho) + \cot(\beta)] \sin(\alpha + \phi + \beta)}{2 \sin(\eta)}, &
        N_Q      = \dfrac{\sin(\alpha + \phi + \beta)}{\sin(\eta)} \\
        N_c      = \dfrac{\cos(\phi)}{\sin(\beta)\sin(\eta)}, &
        N_{c_a}  = \dfrac{-\cos(\rho + \phi + \beta)}{\sin(\rho)\sin(\eta)}
    \end{array}
\end{align}
where the soil properties are the cohesion $c$, internal angle of friction $\phi$, moist unit weight $\gamma$, adhesion $c_a$, and soil-tool friction angle $\delta$.
The surcharge force is defined as $Q$ and the remaining parameters describe the geometry with $\eta = \delta + \rho +\phi + \beta$ defined for notational convenience.

Existing soil property estimation methods for earthmoving applications are aimed at excavation or scooping operations and assume the measurement of multiple (3-8) full dig cycles in the same spatial region prior to producing an estimate \cite{Luengo1998,Tan2005,Yu2023}.
This is not adequate for bulldozing operations where soil-tool interaction takes place over a large area sometimes with only partial overlap between cuts.
By combining a fast physics infused neural network (PINN) approach to soil property estimation with a elevation mapping system, estimates can be tracked spatially and fused over time via Bayesian updates, the details of which are laid out in Section \ref{sec:methods}.

\subsection{Mapping for Autonomous Earthmoving}
Building and maintaining a detailed persistent representation of the environmental state can be beneficial at both control and planning levels, enabling consideration of approaching terrain shape in machine blade control and allowing planning over large regions outside of the machine's immediate field of view.
To represent the topography of terrain for earthmoving, often a raster format is used, where spatial data in the horizontal plane is discretized and elevation at each location is represented continuously.
This is a fairly common approach to mapping in robotics \cite{Cremean2005,Fankhauser2014,Miki2022} that extends simple occupancy grid representations, and these maps are often referred to as elevation maps, heightmaps, heightfields, digital elevation models (DEMs), digital surface models (DSMs), or gridmaps.
For clarity, we will adopt the heightmap nomenclature and use gridmaps when describing the extension to additional layers.
Mononen et. al. use a Bayesian mapping approach to produce online updated heightmaps from range measurements obtained from a forward-facing LiDAR sensor mounted to a bulldozer chassis \cite{Mononen2022a}.
A smooth spline is fit to the approaching surface and adjusted based on blade load and track slip to produce the target blade trajectory.

In place of or in addition to mapping the terrain with vehicle mounted depth sensors, the elevation of the tractor can be used to update the terrain height under the machine tracks, as is done on some commercial systems \cite{Komatsu2022}.
Further, given knowledge of the trajectory that the blade takes, intersections between the blade and map can be determined and used to move material within the map \cite{DAdamo2023,Brewster2018}, i.e. proprioceptive \cite{Jud2019} or blind mapping.
By modeling material movement, real-time estimation of the terrain changes in areas not visible from machine mounted sensors due to obscuration from the blade and terrain is enabled.
This approach can facilitate tracking earthmoving efficiency via estimation of material cut and fill volumes.
These maps provide a natural representation on which to plan earthmoving sequences.

D'Adamo describes a method for estimating the maximum pushable volume for a bulldozer bulk-earthmoving task where geometric, tractive, and engine limits are determined from blade and track geometry, known fixed soil properties, and engine torque curves \cite{DAdamo2023}.
These limits are used to help compensate for vertical blade localization errors when mapping terrain, but it is implied that a similar model is used by the Caterpillar SATS planner when sequencing push actions.

A heightmap can be extended to include additional information by simply adding additional raster layers to represent other qualities relevant to navigation and earthmoving.
We will refer to these extended representations as gridmaps.
This may include tracking of loose soil separately from undisturbed terrain \cite{DAdamo2023}, categorical information produced by a semantic segmentation system \cite{Guan2022,Erni2023}, traversability scores \cite{Miki2022}, or even material properties \cite{Chen2024}.
Guan et. al. leverage gridmaps to build a traversability estimation pipeline that combines geometric features with semantic information to enable improved planning for excavator navigation at a worksite \cite{Guan2022}.

\section{Methods}\label{sec:methods}
In our prior work \cite{Wagner2023}, the core physics infused neural network (PINN)-based soil property estimation system was developed, and simulation-based experiments demonstrated the merits of this approach for the task of rapidly estimating soil properties and strength using the bulldozer itself as a sensor. 
While promising, this approach has limited direct applicability as it relies on the assumption of flat terrain in order to obtain the geometric parameters of the fundamental equation of earthmoving (FEE) $(\alpha, \rho, d, w)$ and it presumes knowledge of privileged information such as the surcharge force $Q$ obtained using a virtual soil mass sensor, of which there is no analog for on a real bulldozer.
Additionally, all prior experiments were performed using the simulation of a simple bladed vehicle model lacking a drive-train and constrained to move along a vertical plane along a flat, level surface.

To enable soil property estimation in more realistic earthmoving scenarios and to enable future soil-aware earthmoving planning, a number of significant advancements are required.
The major contributions of this work therefore include:
\begin{itemize}
    \item Development and demonstration of a system performing soil property estimation in-situ on an advanced simulation of a compact track loader (CTL) with articulated bulldozer blade performing earthmoving operations.
    \item Extension of the physics infused neural network approach to estimate parameter uncertainties without requiring knowledge of true soil properties during training
    \item Development of a swept-volume based approach to heightmap deformation and concurrent extraction of FEE parameters, including surcharge, when model assumptions are met
    \item Development of a Bayesian estimation method for merging multiple soil property estimates and gridmap-based approach for tracking these estimates spatially
\end{itemize}

\subsection{Physics Infused Neural Network for In-Situ Soil Property Estimation} \label{sec:PINN}

A key component of the soil property estimation approach taken here is to bake-in knowledge of soil-tool interaction physics to the network architecture in the form of Equation \ref{eq:FEE_v2}, via a physics infused neural network (PINN) \cite{Wagner2023}.
In comparison to the previously developed model, a number of improvements have been made to both the network architecture and training of the system to generalize the approach to uneven terrains, increase accuracy, and facilitate Bayesian mapping including:
\begin{itemize}
    \item Training on a more realistic and diverse dataset
    \item Building in additional physics using error propagation, and leveraging a negative-log likelihood (NLL) loss term to enable estimation of soil property estimate uncertainty $\Sigma_{\Theta_u}$
    \item Expanding estimated soil properties to include $\gamma$ and $c_a$ and improving estimates by making multi-step force predictions throughout a sweep to reduce ambiguity in parameter estimation
\end{itemize}

\begin{figure}[h!]
    \centering
    \includegraphics[width=1.0\linewidth]{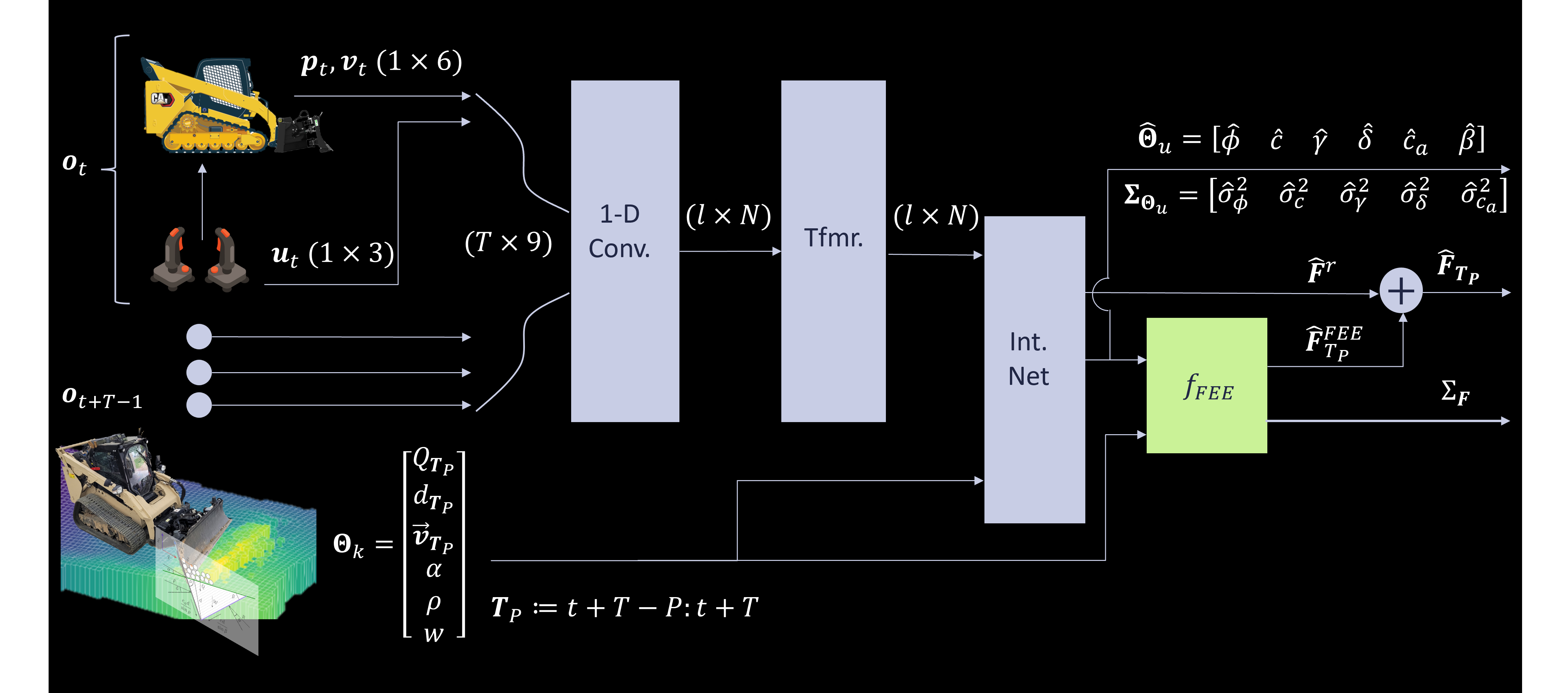}
    \caption{PINN-based soil property estimation system diagram. Inputs flow from the left to produce the estimates on the right. The PINN block is highlighted in in green.}
    \label{fig:network}
\end{figure}

\begin{figure}[h!]
    \centering
    \includegraphics[width=0.85\linewidth]{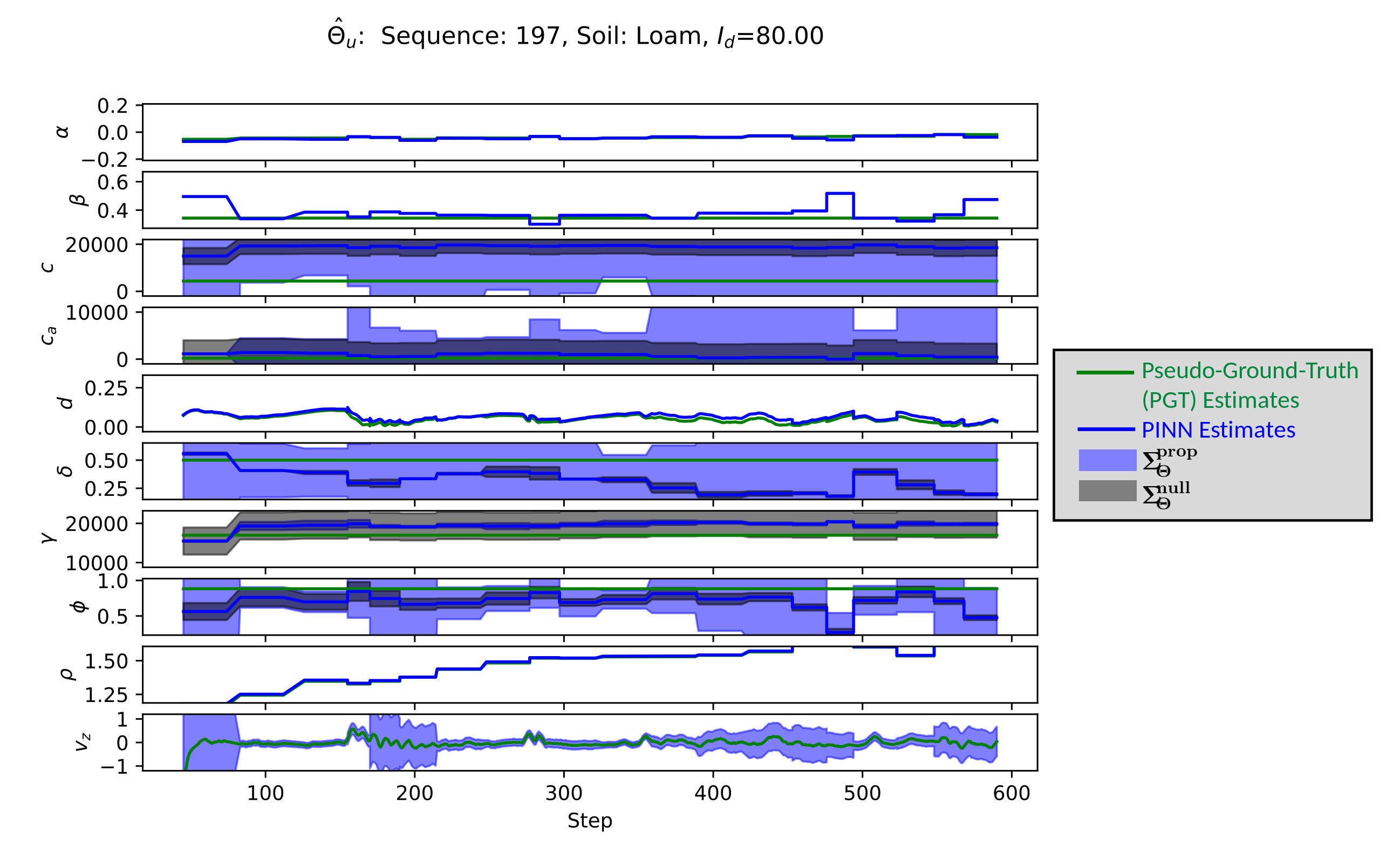}
    \includegraphics[width=0.85\linewidth]{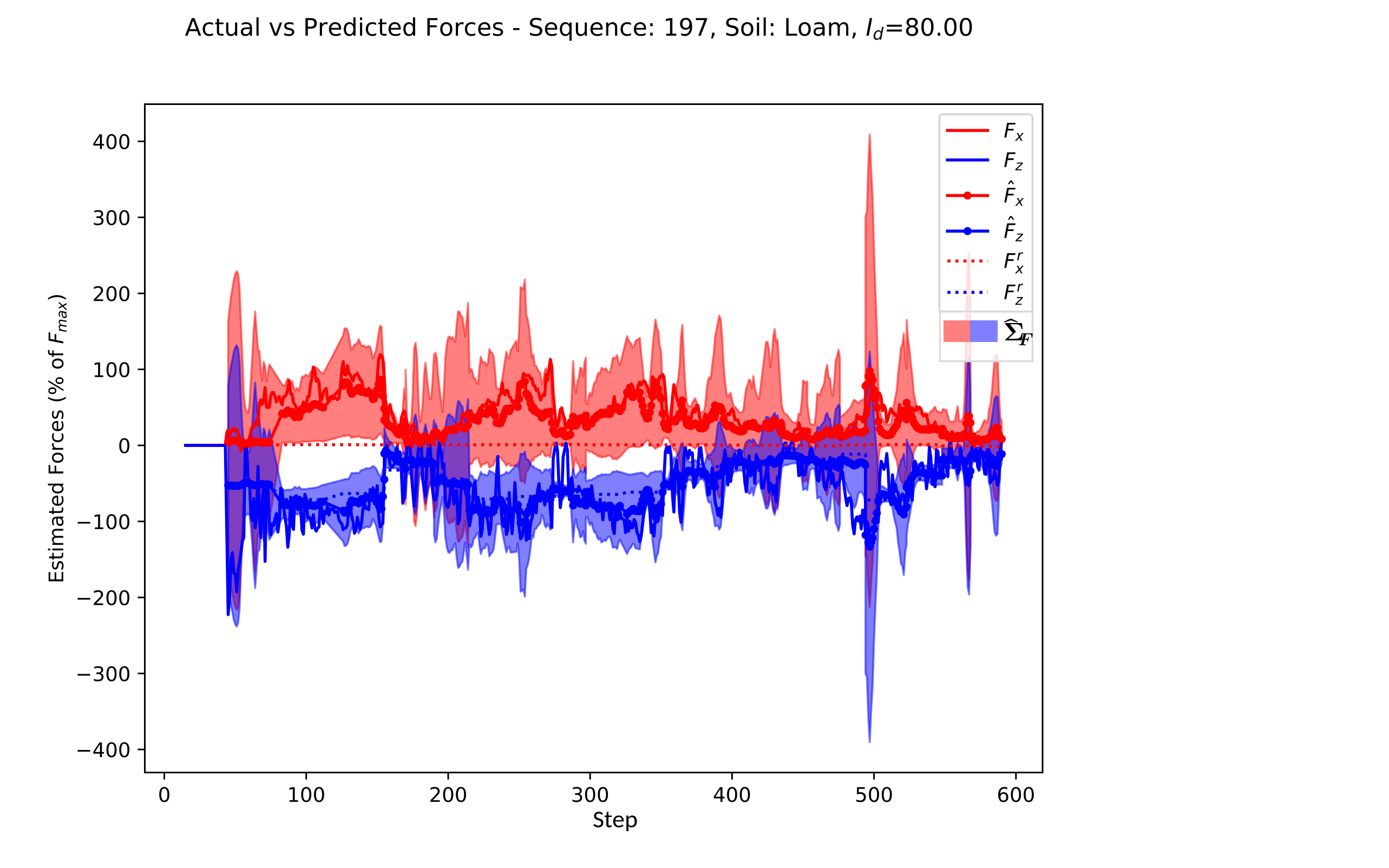}
    \vskip 1em
    \caption{PINN predictions for a single rollout in $I_d=\SI{80}{\percent}$ loam soil. (top) Predictions of soil properties $\hat{\Theta}_u$ (blue line), estimated parameter uncertainties $\Sigma_\Theta$ (blue shading), and nullspace projected uncertainties $\Sigma_\Theta^n$ (gray shading) overlaid on PGT soil properties $\Theta_u$ (green line). (bottom) Corresponding force predictions for the same rollout.}
    \label{fig:rollout}
\end{figure}

As outlined in Figure \ref{fig:network}, the network ingests a history of vehicle states and actions $\b{o}_{[t:t+T]}$ along with a subset of known FEE parameters $\b{\Theta}_k = [\rho, \alpha, w, \b{d}_{\b{T}_P}, \b{Q}_{\b{T}_P}, \b{v}_{\b{T}_P}]$ over the prediction horizon $\b{T}_P \allowbreak =\allowbreak [t+T-P:\allowbreak t+t]$.
It produces, as intermediate outputs, an estimate of the unknown soil properties $\hat{\Theta}_u = [\hat{c},\hat{\phi},\hat{c}_a,\hat{\delta},\hat{\gamma}]$ and a predicted variance for each of these estimates $\Sigma_{\Theta_u} = \text{diag}[\hat{\sigma}_c^2, \hat{\sigma}_\phi^2, \hat{\sigma}_{c_a}^2, \hat{\sigma}_\delta^2]$, as well as predicted variances for a subset of the known parameters.
These estimates are then provided to the embedded FEE block to produce the predicted force $\b{F}_{\b{T}_P}^\text{FEE}$ and propagated uncertainty $\b{\Sigma}_F$.
A residual force $F^r$ is also directly produced by the network and summed with $\b{F}_{\b{T}_P}^\text{FEE}$ to produce the final force prediction $\b{F}_{\b{T}_P}$.
This residual force serves the purpose of compensating for forces not well captured by the FEE model such as those caused by penetration and blade acceleration.
An example rollout of both the estimated parameters and the predicted forces is shown in Figure \ref{fig:rollout}.

For brevity, some of the details of the loss function definition are omitted here; similar formulations can be found in our prior work \cite{Wagner2023}.
The two major changes made here are the averaging of the loss over the multi-step prediction horizon $\b{T}_P$\footnote{To reduce notational clutter, the subscript $\b{T}_P$ indicating a quantity is a vector over the time window has been removed, but the bold facing remains where required.} and the inclusion of an additional negative log-likelihood (NLL) loss term
\begin{equation} \label{eq:NLL}
        \mathcal{L}_\text{NLL}(\b{F}, \hat{\b{F}}, \b{\Sigma}_{\b{F}}) = \frac{1}{2} \left[ (\b{F} - \hat{\b{F}}) \b{\Sigma}_{\b{F}}^{-1} (\b{F} - \hat{\b{F}})^\top + \ln (\det(\b{\Sigma}_{\b{F}})) \right]
\end{equation}
which is derived by formulating the optimization of the distribution $F \sim \mathcal{N}(F(\Theta), \Sigma_F(\Theta,\Sigma_\Theta)$ as a minimum NLL problem.
The challenge here is that a standard NLL loss will only enable determination of uncertainties for the quantity being regressed upon, i.e. the cutting force $F$, but we are interested in obtaining parameter estimate uncertainties $\Sigma_{\Theta}$ not $\Sigma_{F}$.
Since the true parameter values are not assumed to be known, the trick employed here is to require the network to produce $\Sigma_{\Theta}$ and propagate that uncertainty through the FEE Equation \ref{eq:FEE_v2} to obtain $\Sigma_{F}$ as a function of $\Theta$ and $\Sigma_\Theta$.
This is accomplished here using a local expansion method which is suitable when uncertainties are relatively small and where the function is not highly non-linear.
The first order Taylor series expansion of $f_\text{FEE}$ around the estimate $\hat{\Theta}$ is
\begin{align}
    F \approx f_\text{FEE}(\hat{\Theta}) + J(\Theta - \hat{\Theta})
\end{align}
where $J_{ij} = \frac{\partial F_i^\text{FEE}}{\partial \Theta_j}$ is the Jacobian.
The propagated force covariance matrix can then be defined as
\begin{equation}
    \Sigma_F = J \Sigma_\theta J^T
\end{equation}
In practice, the soil parameter uncertainty is not estimated directly by the network.
Instead, the network produces $S_{\Theta_u} = \ln( \Sigma_{\Theta_u})$ resulting in the modified error propagation
\begin{equation}
    \Sigma_F = J \exp(S_\Theta) J^T
\end{equation}
This change is suggested to avoid the potential division by zero in Equation \ref{eq:NLL}, because it is more numerically stable and doesn't require resolving network outputs to the positive domain \cite{Kendall2017a}. 

Overall, this error propagation approach enables us to train the network to estimate parameter uncertainties via back-propagation through Equation \ref{eq:NLL} without requiring knowledge of the true $\Theta_u$.
Uncertainties obtained in this manner are aleatoric, and reflect uncertainty of the underlying noisy force measurements.
This will not capture epistemic uncertainty of the parameter estimates which may arise during inference when, for example, testing on a soil that was not present in the training dataset.

During training it was observed that some parameter estimates were overly confident.
Upon closer inspection, it was observed that under certain conditions, the Jacobian $\b{J}$ is not well conditioned, i.e. has very small singular values.
When this occurs then the operator has a non-zero null space, meaning that changes in parameters and parameter variances on this null space do not affect changes in $\b{F}$ and $\b{\Sigma}_F$.
The PINN relies upon kinematic observations and the action history $\b{o}_{[t:t+T]}$, not the force, to make the parameter estimates, but the force is the only supervisory signal used to train the network.
Therefore, when the Jacobian is deficient, during training the supervision of the parameters that lie within the null space collapses, and during inference the network is unlikely to produce reasonable estimates of those parameters under these conditions.





Similar problems have been observed in other soil property estimation efforts in which non-linear optimization methods are used to solve the inverse terramechanics problem, i.e. estimate $\Theta$ from $F$.
Luengo et al. perform a sensitivity analysis using true parameters as a baseline and individually perturbing the each value to observe the corresponding change in predicted force \cite{Luengo1998}.
For the soil considered and assuming a fixed dig angle, it was found that the cohesion $c$ was the least sensitive to perturbation out of the considered parameters $[\phi,\delta,\beta,c]$ where a 50\% change in $c$ resulted only in a 25\% change in the force. 
Alternatively, Tan et al. investigate the sensitivity of the parameter estimation when the measured force is perturbed, and find that $\phi$ is impacted minimally, while $\gamma$ is much more easily influenced, e.g. changing by a factor of 23\% for a 5\% perturbation of the force \cite{Tan2005}.

While these analyses are somewhat specific to the optimization methods used, they indicate that certain parameters may be more difficult to estimate accurately due to the structure of the FEE.
This occurs acutely when parameters lie within the null space of the Jacobian.
In the mapping task outlined here, downstream of estimation, soil-property estimates are fused together in a Bayesian manner.
In order to avoid the fused estimates from collapsing erroneously due to an artificially confident individual estimate, it is preferable to inflate $\Sigma_\Theta$ so that if $\b{J}$ is well conditioned on a subsequent pass, the likely more accurate parameter estimates will be given more weight.
A simple way of inflating the covariances when $\b{J}$ is singular is to project a fixed covariance into the nullspace
\begin{equation}
    \Sigma_\Theta^n = (I - \b{J}^\dag \b{J})\Sigma_\Theta^a
\end{equation}
where $\Sigma_\Theta^n$ is the projected nullspace covariance, $\Sigma_\Theta^a$ is a fixed value referred to here as the ambiguity covariance, and $\dag$ refers to the pseudo-inverse operation.
The projected nullspace covariance is then combined with the estimated covariance as $\Sigma_{\Theta_u} = \min(\Sigma_{\Theta_u},\Sigma_{\Theta_u}^n)$ to ensure that the uncertainties do not become erroneously too small.






Depending on the data and the number of timesteps for which a set of predicted parameters is assumed constant, i.e. $P$, $\b{J}_{T_P}$ may or may not have a nullspace.
However, it almost always has some very dominant singular values and some much smaller, near zero ones. 
When performing the pseudo-inverse, both absolute $a_\text{tol}$ and relative tolerance $r_\text{tol}$ can be defined to set the threshold for selecting which singular values and corresponding singular vectors should be removed to produce a low-rank approximation of the inverse.
In the current implementation, these values are set to $a_\text{tol} = 1,500$ and $r_\text{tol} = 0.05$, however, the resulting $\b{\Sigma}_\Theta^n$ is not particularly sensitive to these values.
This is related to the magnitude of the parameter values and methods for scaling this projection or modifying the underlying optimizer may improve overall performance \cite{thuerey2021pbdl}. 

The overall loss function is expressed as
\begin{align}
    \mathcal{L}(\b{F}, \hat{\b{F}}, \hat{\b{\Theta}}, \b{F}^r, \frac{\partial N_{\gamma}}{\partial \beta},\b{\Sigma}_{\b{F}}) = 
    &\lambda_{\b{F}} \mathcal{L}_\text{wMAE} (\b{F}, \hat{\b{F}}, w_{xz}) + 
    \lambda_{\text{NLL}}  \mathcal{L}_\text{NLL}^{\Sigma_{\Theta} \leftharpoonup}(\b{F}, \hat{\b{F}}, \b{\Sigma}_{\b{F}}) + \nonumber\\
    &\lambda_{\partial N_{\gamma}/\partial \beta} \mathcal{L}_\text{MAE}^{\beta\leftharpoonup} (\b{0}, \frac{\partial N_{\gamma}}{\partial \beta}) + 
    \sum_{\b{\hat{\theta}} \in \Theta^\text{rng}}\left[\lambda_{\theta} \mathcal{L}_\text{ReLU} (\b{\hat{\theta}}, l^\theta)\right] + \nonumber \\
    &\sum_{\hat{\theta}^\text{res} \in \Theta^\text{res}}\left[\lambda_{\theta^\text{res}} \mathcal{L}_\text{MSE}(0, \hat{\theta}^\text{res}) \right] + 
    \lambda_{F^r} \mathcal{L}_\text{wMAE}(\b{0}, F^r, w_{xz})
    \label{eq:Loss_w_unc}
\end{align}
The first term a weighted mean average error (MAE) loss based on the predicted force error and is the primary supervised loss component.
The second term is the supervised NLL loss from Equation \ref{eq:NLL}, but gradients wrt. this loss are only back-propagated through $\Sigma_\Theta$ as indicated by the superscript $\mathcal{L}^{\Sigma_{\Theta}\leftharpoonup}$.
The remaining four terms are regularization losses whose purpose is to enable consistent estimation of $\beta$, enforce estimated parameters to lie with physically meaningful ranges $l^\theta$, allow for small residual corrections of the FEE parameters extracted via the mapping process $\hat{\theta}_u = \theta_k + \hat{\theta}^\text{res}$ for some subset $\Theta^\text{res} \in \Theta_k$, and ensure that the residual force $F^r$ remains small to compensate for mismatch between the FEE and the true physics.
The $\lambda$ terms refer to weights of each loss component and the loss functions containing the argument $w_{xz}=[1.0, 0.5]$ are used to weight the $x$ and $z$ components separately.
It is also important to note that the parameter uncertainty values produced by the network are also enforced to lie within a reasonable range using the ReLU regularization $\Sigma_\Theta \in \Theta^\text{rng}$.
When any of the following parameters exceed their clipping limits $[\eta, \beta, \rho]$ or if $\hat{d}=d+\hat{d}^\text{res}<0$, the FEE is not valid and therefore all losses except for $F^r$ regularization are invalidated.
This loss is averaged over the prediction horizon $P$ helping to reduce the size of the Jacobian nullspace, and mini-batches of size 100 are used to help stabilize training.

To further improve estimation of the soil failure angle, a nominal value of $\beta$ as a function of $\phi$ is obtained to enable the network to estimate the residual angle $\beta^\text{res}$ as opposed to determining it directly.
This expression is derived by considering a frictionless $\delta=\SI{0}{\degree}$ vertical blade $\rho=\SI{90}{\degree}$ cutting through a flat terrain $\alpha=\SI{0}{\degree}$ comprised of a cohesionless soil $c=0, c_a=0$ \cite{miedema_2019}, yielding
\begin{equation}
    \beta = \frac{\pi}{4}-\frac{1}{2}\phi
\end{equation}

\subsection{Terrain Mapping} \label{sec:EM}
To enable use of this soil property estimation approach on a realistic compact track loader (CTL) simulation performing earthmoving operations on an uneven terrain while articulating the blade to vary its roll, pitch, yaw, and height, a method is devised to obtain the required network input parameters $\Theta_k$ via tracking of blade movement through a terrain map. 
The basic idea is to represent the terrain as a heightmap and then track the movement of the blade as it intersects the heightmap in order to both update the map by displacing soil and to and extract approximate FEE parameters $\Theta_k$ from the geometric relationship between the blade and map.
Algorithm \ref{alg:EM_SV} outlines this procedure and Algorithms \ref{alg:EM_FEE_extraction}-\ref{alg:EM_Soil_Erosion} provide more of the details.
The open source Elevation Mapping cupy software provides an implementation of a GPU accelerated heightmapping approach used in legged robotics applications and offers support for custom semantic mapping and plugins \cite{Miki2022,Erni2023}.
This package acts as foundation that is modified and extended to enable the desired earthmoving applications.

In addition to tracking the terrain height, a gridmap layer containing disturbed/loose soil is also maintained, similar to the method proposed by D'Adamo \cite{DAdamo2023}, enabling computation of surcharge volume $V^Q$ and application of heightmap erosion to loose soil alone.
Given the focus of this work is on using mapping as a means to obtain $\Theta_k$, some additional factors must be considered to enable accurate soil property estimation.
In derivation of the FEE, a number of assumptions are made that must be satisfied including: the blade is in contact with the soil, the soil is failing as a result of the applied force, the forces are balanced i.e. in static equilibrium, the blade is flat, and the cut geometry is uniform along the width of the blade.
By carefully designing the methods used to determine if the blade is in contact with the heightmap and how soil is displaced when contact occurs, these requirements can be met. 

\begin{figure}[h]
    \centering
    \includegraphics[width=0.5\linewidth]{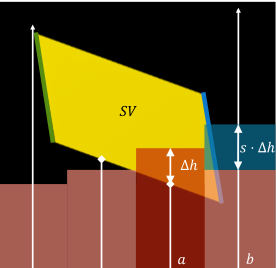}
    \caption{A profile view of the swept volume $SV$ intersecting the elevation map $M$. The ray casting procedure is illustrated with the white lines emanating from the bottom of the cells with intersections marked with white diamonds. The displaced terrain height $\Delta h$ is determined from difference between the cell height and intersection height and the volume is then swelled and deposited in cell $b$.}
    \label{fig:swept_volume_profile}
\end{figure}

A volume, represented as a polygonal mesh, is swept out by the movement of the blade once a minimum translational or rotational displacement has occurred, and subsequently this swept volume is checked for intersection with rays emanating from the center of each heightmap cell, see Figure \ref{fig:swept_volume_profile}.
This differs from the blind mapping terrain update described by D'Adamo in which a fixed solid volume representing the Ground Engaging Tool (GET) geometry is intersected with the heightmap at a fixed rate \cite{DAdamo2023}.
Due to the discretization of the terrain in the horizontal plane implicit with a heightmap representation, it is possible for intersections between the map and a fixed GET volume to be missed, depending on the GET velocity, GET thickness, and update rate, although this was not reported.
Additionally, it isn't obvious how such an approach could be utilized to ensure the FEE assumptions are being met.
For example, using this solid volume intersection approach, it would be difficult to automatically differentiate between the following states: active cutting, a stalled blade still exerting force on the soil, and a blade merely making contact with the terrain but not exerting a force on the soil.
By instead considering the intersection between a GET swept volume and the heightmap, any detected intersections necessarily imply contact and soil failure i.e. the active cutting scenario, thereby satisfying two of the FEE assumptions.

In simulation of soil-tool interaction, Holz et. al. similarly rely upon swept volumes for soil displacement, but instead leverage a discretely sampled signed distance function to represent the terrain surface more continuously \cite{Holz2013}.
This work also hints at a method for satisfying the FEE assumption of a flat blade with uniform cut geometry, by breaking the cutting surface up into multiple thin slices where these assumptions are close to valid.
The cutting force for each slice can then be computed independently using the known soil properties and summed across the blade to produce an aggregate force and torque applied to the blade by the soil.

The objective here is the inverse, to obtain soil property estimates from observation of cutting force, but this slicing concept can be leveraged to back out $\Theta_k$ as laid out in Algorithm \ref{alg:EM_FEE_extraction}.
For each cell intersected by the swept volume, points on the terrain surface along the direction of GET translation are obtained using a Digital Differential Analyzer (DDA) line rasterization algorithm.
During this process, the volume of surcharge lying in front of the blade $V^Q$ is obtained.
A line is then fit to these points using weighted least squares to better fit points close to the blade, and the geometric parameters of the FEE for the slice $(\alpha, \rho, d)$ are obtained.

While it may be possible to develop a soil property estimation algorithm that directly consumes FEE parameters for each slice, for simplicity we have chosen to obtain a single parameterization of the FEE for each sweep.
The cutting force, Equation \ref{eq:FEE_v2}, is strongly sensitive to $d$ and therefore slice FEE parameters are averaged together with a weight determined by $d$.
Averaging these values is somewhat crude, as Equation \ref{eq:FEE_v2} is non-linear, but in practice it has been sufficient.
Although not shown in Algorithm \ref{alg:EM_FEE_extraction}, to enable the model to compensate for poor extracted $\Theta_k$, average goodness of fit metrics the line fit and parameter variances are computed and provided as inputs to the integration network.
To obtain the blade width $w$ each intersection point is projected onto the blade along the translation direction $\hat{t}$ and the range of the points along the direction perpendicular to the translation $\perp \hat{t}$ is determined to be the equivalent blade width.
During partial contact between the blade and the ground $w$ will be less than the full blade width and during cuts where the blade is yawed to one side the width is reduces in proportion to the cosine of the yaw angle.
This entire FEE parameter approximation and extraction procedure is necessary, as it is important for the system being designed to support uneven terrains along with blade roll and yaw because the goal is to enable soil property estimation during performance of typical earthmoving tasks when these configurations will be encountered.

\begin{figure}[h]
    \centering
    \includegraphics[width=0.9\linewidth]{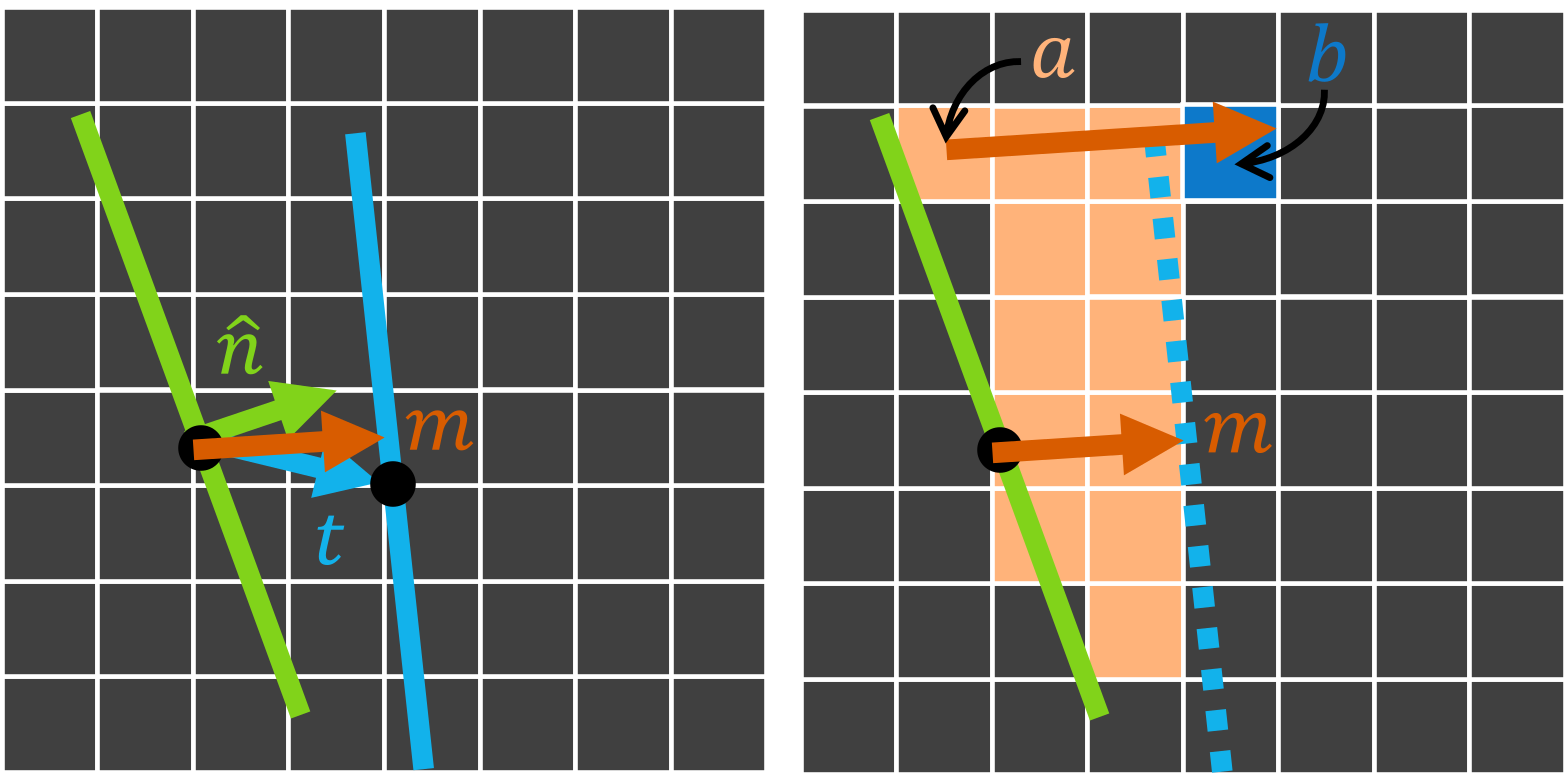}
    \caption{(Left) depiction of the method used for determining the displaced cell deposit direction $\hat{m}$ from the blade normal $\hat{n}$ and translation vector $t$. Note that $\hat{m}=m/||m||$. (Right) Once a deposit direction is selected, a DDA algorithm is used to search along the line beginning at the center of an intersected cell $a$ with the orientation of $\hat{m}$ to find the first cell $b$ that is not intersected in the sweep where the soil is deposited.}
    \label{fig:dda_depost}
\end{figure}

Once $\Theta_k$ has been obtained, updating the map as a result of the sweep can proceed and is described in Algorithm \ref{alg:EM_Cut_Deposit}.
The physics of terrain deformation from soil-tool interaction are very complex, and while the Vortex Studio simulation used for all of the experiments presented here captures some of this behavior, it is still an approximation of reality.
The methods described here are further simplified and rely upon a number of heuristics, but have proven adequate for the application.
For simplicity, it is assumed that when soil is displaced from one cell $a$ it is deposited entirely in another cell $b$.
To determine which cell should receive the displaced material, a DDA rasterization starting at the cell $a$ is performed along the direction $\hat{m}$ to find the first cell lying outside of the swept volume.
The search direction $\hat{m}$ is defined to be between the starting blade normal $\hat{n}$ and blade center translation direction $\hat{t}$.
This process is depicted in Figure \ref{fig:dda_depost}.
Both intersected and deposit cell heights and loose soil values are updated, accounting for swell of the undisturbed material assuming a fixed swell ratio $s$.
The height uncertainties of both cells are also updated conservatively by considering a $1 \sigma$ range of possible cell heights given the prior uncertainties and uncertainty of the swept volume height $\sigma_\text{SV}$.
The change in loose soil volume lying in front of the blade from the start of the sweep to the end of the sweep $\Delta V^Q$ is determined by accounting for loose soil not being displaced, loose soil flowing below the blade, and newly disturbed and swelled soil.

The parameter extraction described in Algorithm \ref{alg:EM_FEE_extraction} yields a single set of values $\Theta_k$ for the entirety of the sweep although it takes place over a period of time in which surcharge is accumulating on the blade and the depth of cut is varying.
To produce a denser dataset for training the network with multiple predictions of cutting force per sweep, the depth of cut $d$ and surcharge volume $V^Q$ are interpolated.
If it is assumed that the disturbed soil has a fixed density $\gamma_l$ the surcharge force $\b{Q}$ can then be obtained from the interpolated surcharge volume.
During training, the extracted parameters $\b{\Theta}_k$ are logged along with the observation history $\b{o}$ from the vehicle's sensors.
The ground truth soil properties $\Theta_u$ can also be logged if available from the simulation or from manual testing, but are not required for training the network.

Once the model is trained, these extracted parameters can be fed to the network to produce the soil property estimates $\hat{\Theta}_u$ and estimate uncertainties $\Sigma_{\Theta_u}$.
These estimates can then be used to perform a Bayesian update of the soil properties as described in Algorithm \ref{alg:EM_Update_Soil}.
Given that soil properties are spatially varying and that these transitions can occur abruptly, e.g. from buried objects or soil strata, we conservatively choose to apply the updates to the region of the map corresponding to the horizontal extent of the FEE soil wedge while inflating covariances for cells further from the blade.

As discussed in Section \ref{sec:PINN}, the accuracy of the estimated soil properties will vary based on both the underlying true soil properties and the geometric configuration of the FEE in relation to the Jacobian $J$.
For example, a shallow cut through a region may result in poor estimation of $\phi$ and $c$ as compared to a deeper cut.
By leveraging Bayesian updates to combine soil property estimates from multiple passes across the same spatial region, the ambiguity inherent in the estimates can be reduced and the accuracy of the estimates can be improved while still accommodating for the spatial variance of soil properties.

The final step of updating the map following a sweep is to perform soil erosion.
This step is necessary to produce realistic looking heightmaps and to account for spill around the edges of the blade during a push.
As discussed earlier, when soil is displaced, it is moved from one cell to another, and over the course of multiple sweeps, loose soil accumulates in the the cells lying before the blade edge, producing a high wall of surcharge.
In reality, as soil is displaced it spreads out and flows until reaching an equilibrium slope, the angle of repose, that is dependent on the soil properties of the disturbed soil.

The erosion method used in Algorithm \ref{alg:EM_Soil_Erosion} lines \ref{alg:line-min-Fs}-\ref{alg:line-compute-slip} is largely based on established methods for soil grid erosion \cite{Li1993, Holz2009}.
Soil slips when the shear stress force $\tau^\prime$ caused by the weight of the soil $W$ becomes greater than the shear strength force $s^\prime$ that is based on the cohesion $c_l$ and internal friction angle $\phi_l$ of the loose soil.
A factor of safety $F_s$ can be defined
\begin{equation}
    F_s(\alpha) = \frac{s^\prime}{\tau^\prime} = \frac{c L + W \cos(\alpha) tan(\phi)}{W \sin(\alpha)}
\end{equation}
that indicates slip will occur when $F_s<1$.
The weight $W$ is a function of the grid resolution and loose soil density $\gamma_l$, and $\alpha$ is the soil failure angle whose range depends on the relative height between adjacent cells.
By minimizing $F_s$ over $\alpha$, it can be determined if slip will occur.
A slip velocity can then be obtained using Newton's second law, and Euler integration can be used to compute the soil slip height $h_\text{slip}$.
In practice, the erosion operation is performed multiple times after each sweep with a small timestep $\delta t$ to ensure smooth erosion and avoid over-sized $h_\text{slip}$.

\begin{figure}[h]
    \centering
    \includegraphics[width=0.4\linewidth]{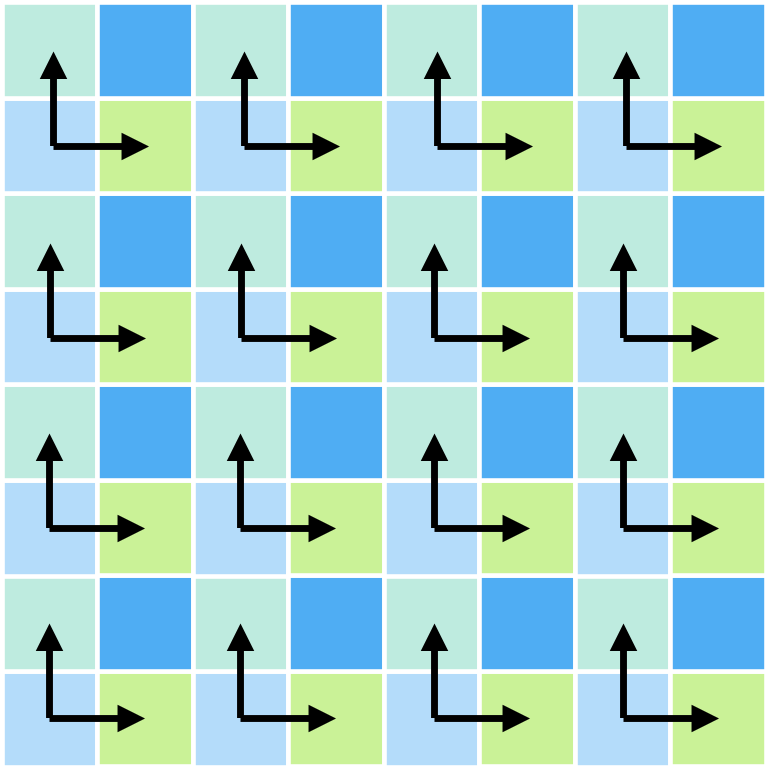}
    \caption{The four different grid colors indicate the four groupings of cells for which slip calculations are performed in parallel and the black arrows depict one of the four possible sets of directions for the cells considered for soil exchange, selected using the deposit direction vector $\hat{m}$.}
    \label{fig:erosion_direction}
\end{figure}

To avoid soil from slipping through the blade, a mask is defined indicating where the swept volume overhangs a given cell.
The height of the bottom of the swept volume at the masked cell centers is also provided to allow for soil to flow from other cells to beneath the blade.
This enables surcharge to slip beneath the blade when it is raised above the compacted soil surface as during a spreading operation.
In order to leverage the GPU to perform these slip calculations in parallel, the slip calculation is applied in a tiled fashion where for each considered cell out of a block of 4 cells, two neighbors at $\SI{90}{\degree}$ apart are considered for soil exchange, see Figure \ref{fig:erosion_direction}.
This is applied in order for each cell in the block and is only performed for a region of interest (ROI) near the blade to reduce computation.
The direction considered for erosion is selected to be in line with the deposit vector $\hat{m}$ so that soil is allowed to erode away from the blade.

\begin{algorithm}
\caption{Update Map with GET Movement} \label{alg:EM_SV}
\begin{algorithmic}[1]
\State $M \gets$ elevation map
\State $\b{T}_{0:1} \gets$ Blade pose history for current sweep
\State $\sigma_{SV} \gets$ Blade height uncertainty during sweep
\State $\gamma_l \gets$ Fixed loose soil moist unit weight
\State $\b{o} \gets$ Observation history
\State $\text{update}, \hat{t}, \hat{n} \gets \Call{CheckMovement}{T_0, T_1}$
\If{$\text{update}$}
    \State $SV \gets \Call{GenerateSweptVolume}{T_0, T_1}$
    \State $\b{A}, \b{\Delta H}, \b{G}, \b{h}^G \gets \Call{IntersectHeightmap}{SV, M}$
    \State $\alpha, \rho, d, w, V^Q_0, \b{P}, \b{X}^{\hat{t}}, \b{Z} \gets \Call{ExtractFEEParams}{M, T_0, \b{A}, \b{\Delta H}, \hat{t}, \hat{n}}$
    \State $\b{B}, \Delta V^Q \gets \Call{CutAndDeposit}{M, \b{A}, \b{\Delta H}, \hat{t}, \hat{n}, \sigma_{SV}}$
    \State $\b{d}, \b{V}^Q \gets \Call{InterpolateFEEParams}{\alpha, \rho, d, \b{T}_{0:1}, V^Q_0}$
    \State $Q \gets \gamma_l \b{V}^Q$
    \LineComment{Collect Known FEE Parameters}
    \State $\b{\Theta}_k = [\alpha, \rho, w, \b{d}, \b{Q}]$
    \If{$ \text{training}$}
        \State $\Call{LogData}{\Theta_k, \b{o}}$
    \Else
        \State $\hat{\Theta}_u, \Sigma_{\Theta_u} \gets \Call{EstimateSoilProperties}{\b{\Theta}_k, \b{o}}$
        \State $\Call{UpdateSoilLayers}{M, \hat{\Theta}_u, \Sigma_{\Theta_u}, \b{A}, \b{P}, x_\text{min}^{\hat{t}}}$
    \EndIf
    \State $\Call{PerformSoilErosion}{M, T_1, \hat{m}, \b{G}, \b{h}^G}$
\EndIf
\end{algorithmic}
\end{algorithm}

\begin{algorithm}
\caption{ExtractFEEParams} \label{alg:EM_FEE_extraction}
\begin{algorithmic}[1]
\State $C_x \gets$ Line fit distance exponential weight 
\State $C_d \gets$ Parameter average depth exponential weight
\Function{ExtractFEEParams}{$M, T_0, \b{A}, \b{\Delta H}, \hat{t}, \hat{n}$}
    \State $\b{P}, \b{X}^{\hat{t}}, \b{Z}, V^Q \gets \Call{GetSurfacePoints}{\b{A},M,\hat{t},\hat{n},T_0}$
    \ForAll{$a,\Delta h,\b{x}^{\hat{t}}, \b{z} \in \b{A}, \b{\Delta H}, \b{X}^{\hat{t}}, \b{Z}$}
        \LineComment{Fit Line to surface points \& Intersect with blade surface \& Extract slice FEE parameters}
        \State $\b{W}_a = \exp(-C_x \b{x}^{\hat{t}})$
        \State $\b{\alpha}[a], \b{\rho}[a], \b{d}[a] \gets \Call{WLSLineFitIntersect}{\b{x}^{\hat{t}}, \b{z}, \b{W}_a, M, \Delta h, \hat{n}, \hat{t}}$
    \EndFor
    \LineComment{Average FEE parameters over slices}
    \State $\b{W}_d \gets \exp(C_d \b{d})$
    \State $\alpha, \rho, d \gets \Call{WAvg}{\b{\alpha}, \b{\rho}, \b{d}, \b{W}_d}$
    \State $\b{P}^{\perp \hat{t}} = \Call{ProjectPointsOnToBladeApprox}{\b{A},\hat{t}}$
    \State $w = \max(\b{P}^{\perp \hat{t}})-\min(\b{P}^{\perp \hat{t}})$
    \State \Return $\alpha, \rho, d, w, V^Q, \b{P}, \b{X}^{\hat{t}}, \b{Z}$
\EndFunction
\end{algorithmic}
\end{algorithm}

\begin{algorithm}
\caption{CutAndDeposit} \label{alg:EM_Cut_Deposit}
\begin{algorithmic}[1]
\State $\lambda_n \gets$ Deposit direction normal weight
\State $s \gets$ Soil swell ratio
\Function{CutAndDeposit}{$M, \b{A}, \b{\Delta H}, \hat{t}, \hat{n}, \sigma_{SV}$}
    \LineComment{Get deposit direction}
   \State $m \gets \lambda_n \hat{n}+(1-\lambda_n) \hat{t}$
   \State $\b{B} \gets \Call{GetDepositCell}{\hat{m}, \b{A}, M}$ 
   \State $\hat{m} \gets \frac{m}{||m||}$
   \State $H := M[\text{elevation}]$
   \State $L := M[\text{loose}]$
   \State $\sigma := M[\sigma]$
    \ForAll{$a, \Delta h, b \in \b{A}, \Delta H, \b{B}$}
        \LineComment{Update intersected cells}
        \State $H[a]^+ \gets H[a] - \Delta h$
        \State $L[a]^+ \gets \max(L[a] - \Delta h, 0)$
        \State $\sigma[a]^+ \gets \max\left(\max(\sigma[a] - \Delta h, 0), \sigma_{SV}\right)$
        \LineComment{Update deposit cells}
        \State $H[b]^+ \gets H[b] + s \Delta h$
        \State $L[b]^+ \gets L[b] + s \Delta h$
        \State $\Delta h_{\max} \gets \Call{Get1SigmaMaxDisplacement}{\sigma_{SV}, \Delta h, \sigma[a], s}$
        \State $\sigma[b]^+ \gets \sqrt{\sigma[b]^2 + \left(\frac{\Delta h_{\max}}{2}\right)^2}$
    \EndFor
    \State $\Delta V_Q \gets \Call{GetDeltaVolume}{H,L,H^+,L^+}$
    \State \Return $\b{B}, \Delta V_Q$
\EndFunction
\end{algorithmic}
\end{algorithm}

\begin{algorithm}
\caption{Update Soil Layers} \label{alg:EM_Update_Soil}
\begin{algorithmic}[1]
\State $C_s \gets$ Soil wedge distance exponential weight
\State $x_\text{min}^{\hat{t}} \gets$ Minimum soil wedge marking distance
\Function{UpdateSoilLayers}{$M, \hat{\Theta}_u, \Sigma_{\Theta_u}, \b{P}, \b{X}^{\hat{t}}$}
    \LineComment{Obtain distance to end of wedge}
    \State $x_\text{max}^{\hat{t}} \gets \frac{d}{\cos(\alpha) (\tan(\alpha+\beta)-\tan(\alpha))}$
    \State $ \b{\Theta_u} := M[\Theta_u]$
    \State $\b{\Sigma}_{\Theta_u} := M[\Sigma_{\Theta_u}]$
    \ForAll{$\b{p}, \b{x}^{\hat{t}} \in \b{P}, \b{X}^{\hat{t}}$}
        \LineComment{Inflate estimated variances}
        \State $\b{v} \gets \b{x}^{\hat{t}} < x_\text{max}^{\hat{t}} \lor \b{x}^{\hat{t}} < x_\text{min}^{\hat{t}}$
        \State $\b{C}_w \gets \exp(C_s\frac{\b{x}^{\hat{t}}}{x_\text{max}^{\hat{t}}})$
        \State $\Sigma_{\Theta_u}^+ \gets \b{C}_w \b{\Sigma}_{\Theta_u}$
        \LineComment{Perform Bayesian update for valid wedge cells}
        \State $\b{\Sigma}_{\Theta_u}^{-} \gets  \b{\Sigma}_{\Theta_u}[\b{v}]$
        \State $\b{\Theta}_u^{-} \gets  \b{\Theta}_u[\b{v}]$
        \State $\b{\Theta}_u[\b{v}] \gets \frac{\Sigma_{\Theta_u}^+ \b{\Theta}_u^{-} +\b{\Sigma}_{\Theta_u}^{-} \Theta_u}{\b{\Sigma}_{\Theta_u}^{-1} + \Sigma_{\Theta_u}^+}$
        \State $\b{\Sigma}_{\Theta_u}[\b{v}] \gets \frac{\Sigma_{\Theta_u}^+ \b{\Sigma}_{\Theta_u}^{-}}{\b{\Sigma}_{\Theta_u}^{-1} + \Sigma_{\Theta_u}^+}$
    \EndFor
    \State \Return
\EndFunction
\end{algorithmic}
\end{algorithm}

\begin{algorithm}
\caption{Perform Soil Erosion} \label{alg:EM_Soil_Erosion}
\begin{algorithmic}[1]
\State $c_l, \phi_l, \gamma_l \gets$ Fixed loose soil properties for erosion
\State $D_\text{ROI} \gets$ Soil erosion ROI dimensions
\State $g \gets$ Heightmap resolution
\State $E \gets \b{0}$ Blade and swept volume mask
\State $Z \gets \b{0}$ Mask heights
\State $\delta t \gets$ Erosion timestep
\Function{PerformSoilErosion}{$M, T_1, \hat{m}, \b{G}, \b{h}^G$}
    \State $\b{R}_a, \b{R}_b \gets \Call{GetROITiles}{T_1, \hat{m}, D_\text{ROI}}$
    \State $H := M[\text{elevation}]$
    \State $L := M[\text{loose}]$
    \State $\sigma := M[\sigma]$
    \LineComment{Fill Erosion Mask}
    \State $E[\b{G}] \gets 1$
    \State $Z[\b{G}] \gets \b{h}^G$
    \ForAll{$\b{r}_a, \b{r}_b \in \b{R}_a \b{R}_b$}
        \State $\b{\Delta h} \gets \min(H[\b{r}_a] - H[\b{r}_b], L[\b{r}_a])$
        \State $F_\text{min}, \alpha^* \gets \Call{MinimizeSlipSafteyFactor}{\b{\Delta h}, c_l, \phi_l, \gamma_l, g}$ \label{alg:line-min-Fs}
        \State $\b{h}_\text{slip} \gets \Call{ComputeSlip}{\alpha^*, \b{\Delta h}, c_l, \phi_l, \gamma_l, g, \delta t}$ \label{alg:line-compute-slip}
        \LineComment{Limit slip if masked}
        \State $\b{e}_a \gets E[\b{r}_a]$
        \State $\b{z}_a \gets Z[\b{r}_a]$
        \State $\b{e}_b \gets E[\b{r}_b]$
        \State $\b{z}_b \gets Z[\b{r}_b]$
        \State $\b{h}_\text{slip}[\b{e}_b] \gets \min(\b{h}_\text{slip}[\b{e}_b], \max(\b{z}_b-H[\b{r}_b],0))$
        \State $\b{h}_\text{slip}[\b{e}_a] \gets 0$
        \LineComment{Update map}
        \State $H[\b{r}_a] \gets H[\b{r}_a] - \b{h}_\text{slip}$
        \State $L[\b{r}_a] \gets H[\b{r}_a] - \b{h}_\text{slip}$
        \State $\sigma[\b{r}_a] \gets \sigma[\b{r}_a]-\b{h}_\text{slip}$
        \State $H[\b{r}_b] \gets H[\b{r}_b] + \b{h}_\text{slip}$
        \State $L[\b{r}_b] \gets H[\b{r}_b] + \b{h}_\text{slip}$
        \State $\sigma[\b{r}_b] \gets \sigma[\b{r}_a]+\b{h}_\text{slip}$
    \EndFor
    \State \Return
\EndFunction
\end{algorithmic}
\end{algorithm}


\section{Results and Discussion} \label{sec:istvs25_results}
\begin{figure}[h]
    \centering
    \includegraphics[width=0.6\linewidth]{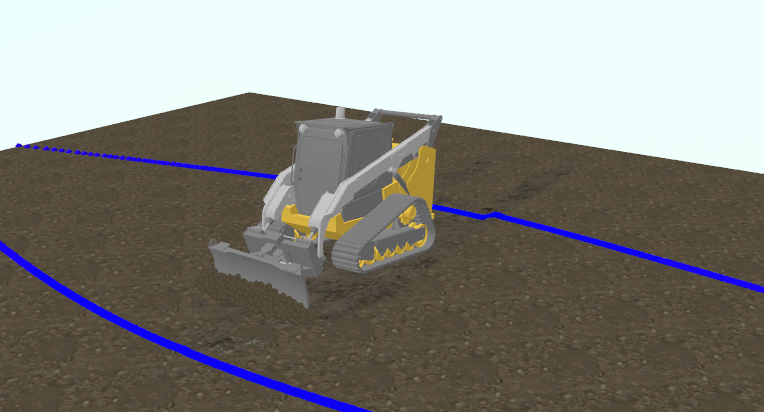}
    \caption{Vortex Studio simulation of Caterpillar 299 D3 XE Compact Track Loader (CTL)}
    \label{fig:CTL_sim}
\end{figure}

All experiments have been performed using a simulated compact track loader (CTL) modeled after a CAT 299 D3 XE outfitted with an articulated bulldozing blade, see Figure \ref{fig:CTL_sim}.
The simulation was constructed by CM Labs within Vortex Studio, which supports  simulation of multi-body dynamics, real-time hybrid soil-tool interaction, low-fidelity hydraulics systems modeling, various virtual sensors \cite{CMLabsSimulations2016}.
Very few specifications of the of the system are available publicly, making development of an accurate simulation challenging.
A mix of physical measurements and photographs were used to produce an approximate CAD model of the system with a focus on proper machine kinematics.
This CAD model, public high-level machine specifications, and hydraulic specifications obtained from the manufacturer informed the simulation development.
Realistic dynamics were prioritized over cosmetic fidelity, but no analysis regarding the accuracy of the model to the real-world system has yet been performed.

The CTL has a higher-center of gravity and a shorter wheelbase than a standard bulldozer, which makes it more sensitive to the terrain underfoot and less stable around the pitch axis in particular.
Oscillations in pitch must be properly compensated for by varying blade height in order to ensure creation of smooth terrain profiles.
The CTL also has a lower drawbar-pull than even small dozers, meaning that the cutting force it can apply to the soil is lower and therefore so is the maximum depth of cut.
Due to these differences, bulldozing with a CTL is a more challenging task than with a conventional bulldozer, but is considered a sufficient analog platform for this work.

Using the Vortex Studio Python API, the simulation can be configured with randomized terrain, parameters for the soil simulation can be specified, commands can be provided to the vehicle actuators and sensor observations can be obtained at a rate of 60 Hz.
The measurements consist of chassis and blade pose and twist, hydraulic cylinder lengths and velocities, track velocities, engine speed, soil cutting force, and pointclouds from front and rear mounted single beam LiDAR scanners.

To enable automatic data collection, cascaded control architecture is developed where at the lower level feed-forward PID velocity controllers are tuned for each arm joint and the tracks.
A combination of individual joint position control and a rudimentary task-space control is implemented using multiple PID controllers to enable tracking of blade height and roll relative to the surface while maintaining a desired pitch joint position and taking into account the loss of manipulability when the lift arm reaches its lowest position.
To keep the vehicle from stalling and still collect a dataset with varying depths of cut, a controller that raises the blade in response to poor tracking of the desired forward velocity is developed.
This approach works reasonably well, but is not optimally tuned and some oscillation occur, particularly upon a substantial step change in the height command.

\begin{figure}[h]
    \centering
    \includegraphics[width=0.9\linewidth]{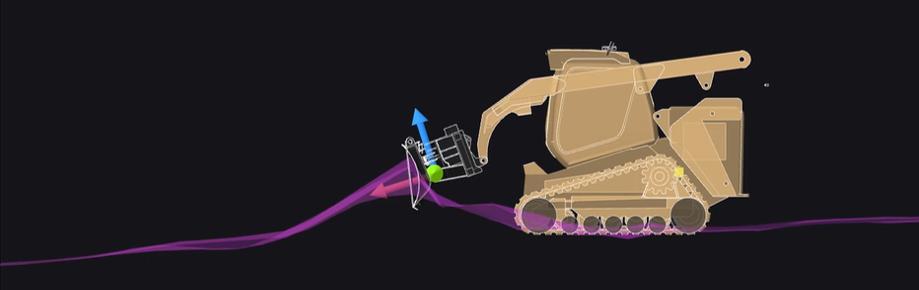}
    \caption{Illustration of how the blind mapping system updates the terrain height in front of the blade during a cut through a berm.}
    \label{fig:mid_cut_hm_profile}
\end{figure}

The terrain mapping system described in Section \ref{sec:EM} is initialized using the starting heightmap from the simulation.
The mapping system is provided with pointcloud measurements from both LiDARS at each timestep as well as the blade pose to be used for blind mapping using blade swept volumes as described in Algorithm \ref{alg:EM_SV}.
See Figure \ref{fig:mid_cut_hm_profile} for a visualization of how the blind mapping enables updating the map in the region directly in front of the blade that is not visible from the front LiDAR.

\begin{figure}[h!]
    \centering
    \includegraphics[width=0.45\linewidth]{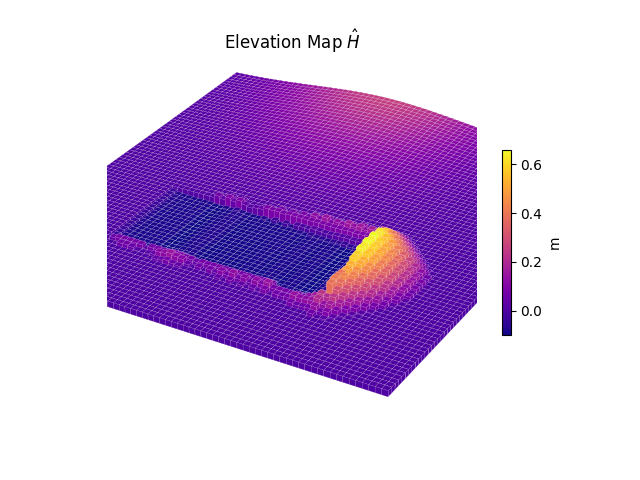}
    \includegraphics[width=0.45\linewidth]{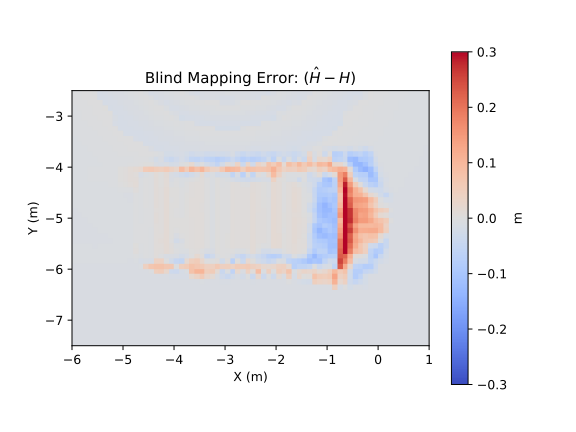}
    \caption{(left) A 3D rendering of a heightmap $\hat{G}$ following a cut when using only blind mapping. (right) The resulting heightmap error.}
    \label{fig:HM_accuracy}
\end{figure}

To validate this approach, tests were performed in which the pointcloud observations were disabled, and only the blade movement was used to update the map from its starting state.
Figure \ref{fig:HM_accuracy} depicts one such experiment.
There are some discrepancies between the estimated height and the true height in two main areas: along the edges of the cut path that was taken by the CTL and where the soil was deposited.
Errors of this nature are expected and are a result of a mismatch between the way that soil particles flow in simulation and the approximation of this process using soil erosion, see Algorithm \ref{alg:EM_Soil_Erosion}.
The primary issue is that the soil erosion parameters are assumed to be fixed $c_l=\SI{25}{\pascal}, \phi_l=\SI{0.26}{\radian}, s=1.2$ when in reality they vary with the undisturbed soil properties.
The reason for this is that outside of simulation these values are not known a priori.
With some tuning, the blind mapping error can be reduced, but this is not necessary to achieve good soil property estimation performance.
Additionally, when pointcloud measurements are enabled, these mapping errors are corrected when in the field of view of the sensor.

\subsection{Soil Property Estimation Accuracy}
A dataset of vehicle observations $\b{o}$, extracted FEE parameters $\b{\Theta}_k$, and pseudo ground truth (PGT) soil properties $\Theta_u$ is collected using the data collection controller and performing a sweep over controller height and forward velocity setpoints.
A total of 268, 10 second long episodes are collected across four pre-tuned soil types (Clay, Loam, Gravel, Sand) where the relative density $I_d$ of the soil is varied from 1-100\%.
Some configuration changes were made to the Vortex Studio soil simulation to ensure more realistic behavior for bulldozing operations and are described in further detail in \cite{Wagner2025FEE}.
To improve the ability of the system to perform in various environments, the terrain heightmap is randomized to small hills and dips using a mixture of Gaussians approach.
Additionally, the blade controller roll and yaw setpoints along with the relative height, forward velocity, and chassis yaw angle setpoints are made to drift throughout an episode using a random walk.

\begin{figure}[htbp]
    \centering
    \begin{subfigure}[b]{0.49\textwidth}
        \includegraphics[width=\textwidth]{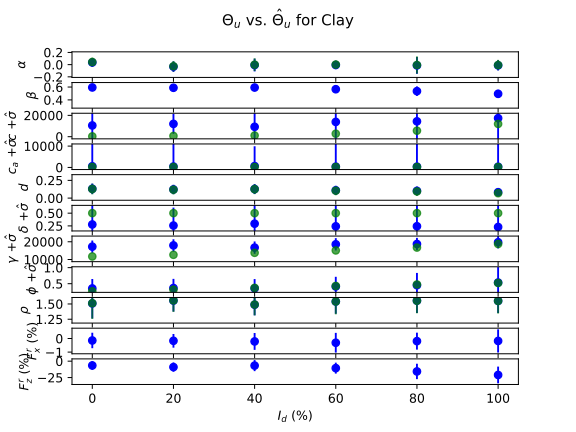}
    \end{subfigure}
    \hfill
    \begin{subfigure}[b]{0.49\textwidth}
        \includegraphics[width=\textwidth]{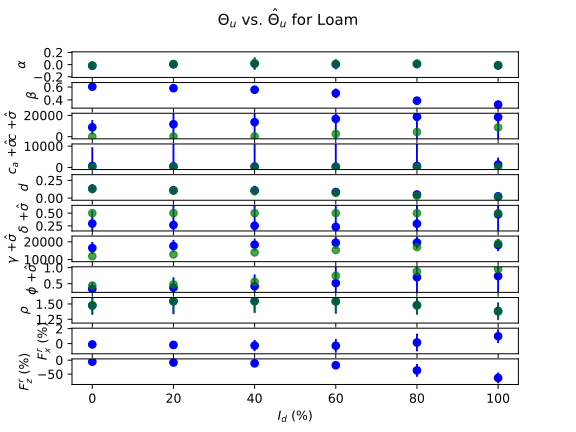}
    \end{subfigure}
    
    
    \begin{subfigure}[b]{0.49\textwidth}
        \includegraphics[width=\textwidth]{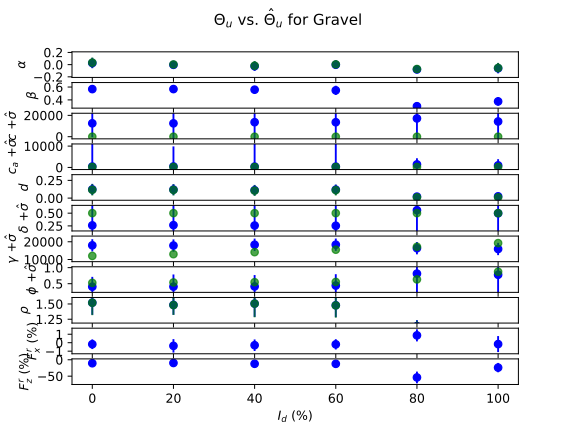}
    \end{subfigure}
    \hfill
    \begin{subfigure}[b]{0.49\textwidth}
        \includegraphics[width=\textwidth]{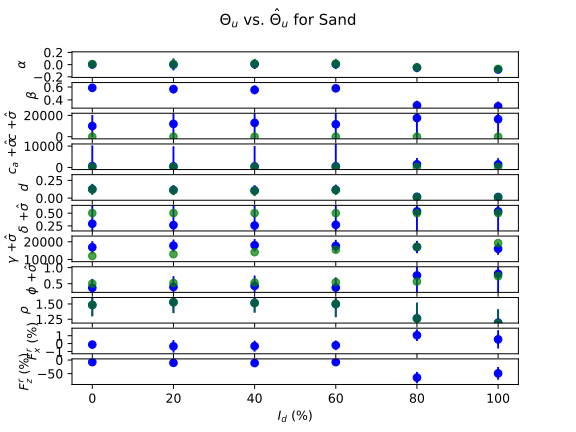}
    \end{subfigure}

    \caption{PINN soil property estimates $\hat{\Theta}_u$ aggregated over the dataset and binned according to the relative density $I_d$ are shown in blue and PGN values $\Theta_u$ are shown in green. The vertical lines indicate the average predicted standard deviation $\Sigma_u^{-1/2}$}
    \label{fig:FEE_params}
\end{figure}

The soil property estimation network is then trained on this dataset.
The model is used to make predictions across the entire dataset and then averaged across values for fixed $\Theta_u$ corresponding to changes in relative density $I_d$.
The averaged soil property predictions are plotted with the PGT values against increasing $I_d$ for each of the four soil types, see Figure \ref{fig:FEE_params}

Before analyzing these results it is important to explain why the soil properties obtained from the simulation $\Theta_u$ are referred to as \emph{pseudo} ground truth (PGT).
Vortex Studio simulates soil cutting forces use a hybrid heightmap-DEM method that utilizes the FEE to determine forces when the shear strength has been overcome by the tool and then spawns particles to enable more realistic dynamic behavior.
A variety of heuristics have also been implemented to improve realism from the perspective of human operators.
So, although the values $\Theta_u$ are obtained directly from the simulation, for the purposes of evaluating the performance of the soil property estimation system, they can only be used as a reference and not absolute ground truth values.

With this in mind, it is useful to compare trend of the estimated parameters $\hat{\Theta}_u$ with relative density $I_d$ to the trend of the PGT properties $\Theta_u$.
Increasing values of $\hat{\phi}, \hat{c}$ and $\hat{\gamma}$ are observed for both of the cohesive soils, and $\phi$ also displays a positive trend for the cohesionless soils.
The friction $\phi$ is estimated particularly well, whereas the estimated density is not terribly accurate.
This aligns with the Jacobian $J$ having very small singular values for $\gamma$ and may be improved with some modifications to the underlying gradient descent method.
In agreeance with our prior results, $\hat{c}$ is consistently significantly larger than the PGT value and thought to be related to additional forces from DEM.
Similarly, for $I_d-[0-60]$, $\hat{\delta}$ is consistently lower, but jumps up to near the PGT value for higher relative densities.
Correspondingly, the average depth of cut across the episode decreases drastically and residual forces $F^r$ increase drastically.
This is reflective of an odd behavior, observed during simulation with hight relative density, where the blade was incapable of penetrating the soil surface even when the entire weight of the vehicle is applied.
While penetration forces are expected to increase with relative density, at slightly lower relative densities the soil can be penetrated and the maximum depth of cut that can be achieved without stalling increases drastically as well.

Without more detailed documentation regarding the Vortex Studio soil simulation, it is difficult to say what the cause of this behavior is, but the important take away is that at higher relative densities $I_d \in \sim [70-100]$, poor soil property estimates are more reflective of issues within the dataset than of the estimation approach.
It is interesting to point out that when this penetration issue occurs, the network produces a large $F^r_z$ to compensate for the non-FEE compatible behavior as intended.
The other important thing to note is that the uncertainty estimates for the parameters also increase in this case which is a desirable feature for the downstream Bayesian mapping.
Taking a closer look at the rollout in Figure \ref{fig:rollout}, the projected force uncertainty is shown to increase when a step change in height is observed at around the 500 step mark.

\subsection{Soil Property Mapping Accuracy}
To qualitatively evaluate the ability of the combined soil property estimation and mapping system to handle spatially varying soil properties, a simple experiment was performed in which a strip of soil with a high $I_d=\SI{80}{\percent}$ is present in a flat terrain with relatively low $I_d=\SI{10}{\percent}$.
The vehicle is then operated by the author to perform a sequence of cuts across the region, where the cut begins on the left in the lower density soil and moves to the right across the strip of higher density soil.
The individual estimated soil properties $\hat{\Theta}_u$ are tracked as separate layers in the map and fused together using Bayesian updates, as outlined in Algorithm \ref{alg:EM_Update_Soil}.

\begin{figure}[h!]
    \centering
    \includegraphics[width=0.75\linewidth]{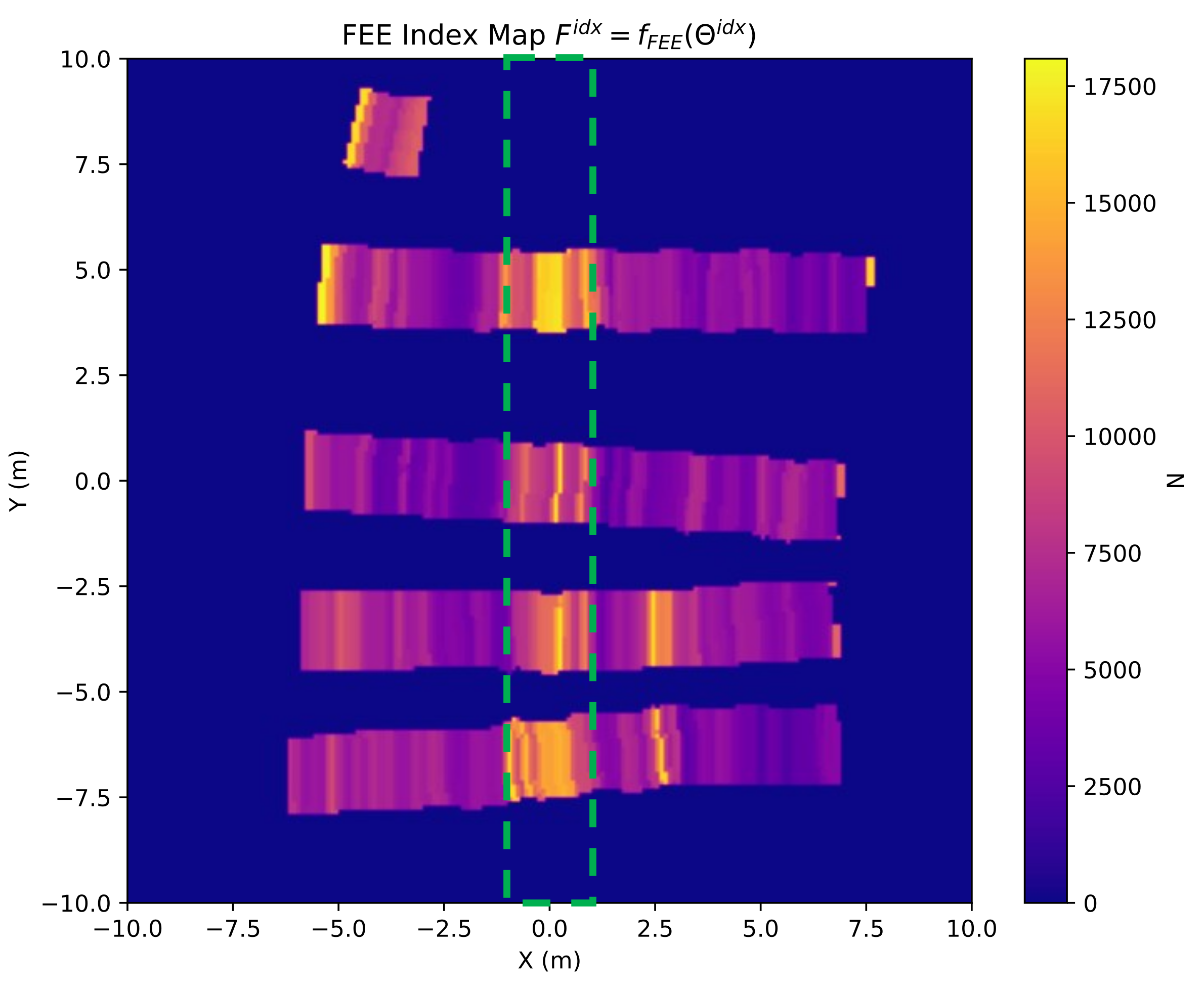}
    \caption{FEE Index map for multi-cut experiment. Clay soil was specified with $I_d=\SI{10}{\percent}$ everywhere except inside the dashed green line where $I_d=\SI{80}{\percent}$.}
    \label{fig:fee_index}
\end{figure}

The fused soil properties from each layer are then used to compute what we define as the FEE index $F^\text{idx} = f_\text{FEE}(\Theta^\text{idx})$, which corresponds the the force that would be required to shear the soil under a specific geometric scenario where $\Theta^\text{idx} = [\alpha=\SI{0}{\degree}, \rho=\SI{80}{\degree}, w=\SI{1.85}{\meter}, d=\SI{0.2}{\meter}, \phi=\bar{\phi}, c=\bar{c}, \delta=\bar{\delta}, c_a=\bar{c_a}, \gamma=\bar{\gamma}]$.
The resulting map for this test is shown in Figure \ref{fig:fee_index}, where the denser soil region is outlined.

The FEE index within the green outline is shown to be substantially higher than the index outside for most of the sweeps.
There are some erroneously higher index regions outside of the higher-density strip.
These errors sometimes occur at the beginning of the push which is a time where the blade is moving somewhat dynamically as the controller adjusts the arm position to arrive at the setpoint.
In this case the assumption of static equillibrium is violated which may explain the inaccuracy to some extent.
The soil property estimation network was designed to partially compensate for this through prediction of the residual force $F^r$ (fixed across the prediction horizon $\b{T}_P$) which is added to the multi-step predicted force $\b{\hat{F}}^\text{FEE}$.
If there exist any oscillations (due to poor controller tuning), then this fixed value residual compensation may provide adequate compensation for the blade acceleration.
Oscillations of the blade both at the beginning of many cuts and when transitioning out of the higher strength soil.

Overall, the results of this test show that our approach to soil property mapping is largely successful.
To arrive at this result, multiple trials were performed and in that process some issues were uncovered.
The primary problem was that in order to perform a a multi-pass test, some minor modifications were made to the controller to enable a person to control the track velocity and chassis angle setpoints.
Additionally, a toggle was implemented to enable the operator to raise the blade when reversing to the following cut start location.
This resulted in some slight differences in the blade pitch angle after lifting, which were not immediately obvious.
While the blind mapping system performed as expected, the soil property estimation system produced poor estimates.
The working theory is that this is because the training dataset did not include any observations with this blade angle and therefore the data was out of distribution.
This can be remedied by improving the blade controller to achieve more consistent, less oscillatory motions and by expanding the training dataset to cover the domain more fully.
However, given the complexity of such a system and the large number of degrees of freedom, it may be challenging to ensure full coverage over the domain.
Therefore this approach might additionally benefit from development of a method for epistimic uncertainty estimation to ensure that when conditions not encountered during training are observed, a high uncertainty is produced so as to not negatively influence the fused map states.

\section{Conclusion}

This work has advanced the state of autonomous earthmoving by extending previous methods for soil property estimation to handle non-flat terrain geometries. A key contribution is the integration of a terrain mapping system, in which a heightmap of the environment is constructed from onboard LiDAR data, and a blind mapping component uses blade kinematics to infer terrain displacement. 
An improved PINN-based model is introduced to estimate a broader set of fundamental earthmoving equation (FEE) soil parameters, along with uncertainty quantification. These estimates are fused into a spatial soil property gridmap using a Bayesian update mechanism, enabling continuous soil-aware terrain modeling during operation.

Initial experiments indicate that this system accurately identifies regions requiring higher interaction forces, showing promise for its application in soil-aware planning and control. Remarkably, despite the increased complexity of operating over uneven terrain, the system achieves better estimation performance than previous approaches limited to flat surfaces. To our knowledge, this is the first system to perform real-time, in-situ soil property mapping.This capability establishes a foundation for intelligent, soil-aware earthmoving systems capable of optimizing cut strategies, avoiding failure conditions, and autonomously responding to subsurface anomalies.

Several avenues remain for future work. The mapping framework could benefit from techniques to mitigate aliasing artifacts, for example by projecting GET motion to apply gradual cell updates. Incorporating a priori soil class knowledge or image-based classification could further constrain the estimation space and improve accuracy \cite{Kasaragod2024}. More sophisticated update strategies, such as inflating covariance with depth or applying observations via Gaussian processes, could enhance robustness in stratified soils.

While this study was conducted in simulation, transitioning to real-world platforms will require new methods for force estimation—either via hydraulic pressure sensing or by leveraging dynamics models to infer force from kinematics. Exploring these directions will be critical for validating the approach outside of controlled environments.
Ultimately, while perfect soil property estimation may be infeasible due to the inverse terramechanics problem, producing estimates that yield accurate force predictions under varying blade configurations may be sufficient for autonomous planning. Where higher fidelity estimates are needed for engineering tasks, improved uncertainty modeling will be necessary.

\printbibliography[heading=bibintoc,title={References}]


\end{document}